\documentclass[11pt]{article}

\usepackage[utf8]{inputenc}

\usepackage{booktabs}
\usepackage{graphicx}
\usepackage{hyperref}
\usepackage{url}
\usepackage{amsmath}
\usepackage{amsfonts}
\usepackage{tabularx}
\usepackage{booktabs}
\usepackage{array}
\usepackage{algorithmic}
\usepackage{gensymb}
\usepackage{color}

\usepackage{geometry}
\usepackage{natbib}
\setcitestyle{numbers,sort&compress}

\title{A Deep Learning Approach for Population Estimation from Satellite Imagery}

\makeatletter
\renewcommand\@date{{%
  \vspace{-\baselineskip}%
  \large\centering
  \begin{tabular}{@{}c@{}}
    Caleb Robinson\\
    \normalsize dcrobins@gatech.edu
  \end{tabular}%
  \quad 
  \begin{tabular}{@{}c@{}}
    Fred Hohman\\
    \normalsize fredhohman@gatech.edu
  \end{tabular}
    \quad 
  \begin{tabular}{@{}c@{}}
    Bistra Dilkina\\
    \normalsize bdilkina@cc.gatech.edu
  \end{tabular}

  \bigskip

  School of Computational Science and Engineering\\Georgia Institute of Technology
}}
\makeatother

\begin{document}

\maketitle

\begin{abstract}
Knowing where people live is a fundamental component of many decision making processes such as urban development, infectious disease containment, evacuation planning, risk management, conservation planning, and more. While bottom-up, survey driven censuses can provide a comprehensive view into the population landscape of a country, they are expensive to realize, are infrequently performed, and only provide population counts over broad areas. Population disaggregation techniques and population projection methods individually address these shortcomings, but also have shortcomings of their own.
To jointly answer the questions of ``where do people live'' and  ``how many people live there,'' we propose a deep learning model for creating high-resolution population estimations from satellite imagery. Specifically, we train convolutional neural networks to predict population in the USA at a $0.01 \degree \times 0.01 \degree$ resolution grid from 1-year composite Landsat imagery. We validate these models in two ways: quantitatively, by comparing our model's grid cell estimates aggregated at a county-level to several US Census county-level population projections, and qualitatively, by directly interpreting the model's predictions in terms of the satellite image inputs. We find that aggregating our model's estimates gives comparable results to the Census county-level population projections and that the predictions made by our model can be directly interpreted, which give it advantages over traditional population disaggregation methods. In general, our model is an example of how machine learning techniques can be an effective tool for extracting information from inherently unstructured, remotely sensed data to provide effective solutions to social problems.
\end{abstract}
 
\section{Introduction}
\label{sec:introduction}

Many countries around the world conduct censuses to gather rich information about their population's size, composition, and demographics. 
While these censuses only happen every 5 to 10 years depending on the country, they are highly important for government policymakers and planners who use population projections to gauge future demand for food, water, energy, and services. 
In the United States sub-national population estimates between census dates are used extensively. 
County level population estimates are used in: ``federal and state funds allocation'', ``denominators for vital rates and per capita time series'', ``survey controls'', ``administrative planning and marketing guidance'', and ``descriptive and analytical studies'', according to Long, 1996~\cite{long1996postcensal}. 
Population projections also impact the economy and may result is large governmental spending. For example, according to the US General Accounting Office, more than ``70 federal programs distribute tens of billions of dollars annually on the basis of population estimates'', and ``[e]ven more money was distributed indirectly on the basis of indicators which used population estimates for denominators or controls''~\cite{long1996postcensal}.
Unfortunately, censuses in many other countries are non-representative due to limited civil registration systems~\cite{balk2006determining}.

Population projection accuracy has also been gaining more attention due to the consequences of long-term health effects such as aging and infectious diseases such as HIV/AIDS.
Traditionally, population predictions rely on the interaction between three factors:  fertility, mortality, and migration\footnote{Public Reference Bureau: \url{http://www.prb.org/Publications/Reports/}}. 
To project population characteristics at a future date, demographers make assumptions about fertility and mortality in a current population and further assume how many people will move into or out of an area before that date, i.e., migration. 
But since population projections carry inherent uncertainty, demographers often times can use previous projections and projection errors to better inform future projections.

Given an administrative area, the spatial distribution of the population in that area can be determined by answering two questions: ``how many people live in the area?", and ``where, specifically, in the area do people live?''. 
These two questions can be cast as the following two problems: population projection, and population disaggregation.
Traditionally, these questions are addressed independently of one another using population projection methods and population disaggregation methods, respectively.
In the population projection task, the goal is to estimate the number of people that live in a particular administrative area based on historical data. 
Methods such as regression models, and non-comprehensive supplemental census surveys (like the American Community Survey) belong to this category. 
In the population disaggregation task, the goal is to distribute a population estimate for a given administrative area within that area, i.e., at a higher spatial resolution than the population estimate was originally made for.  

Our proposed method performs both of these tasks jointly. 
Using recent techniques from deep learning, which has shown remarkable state-of-the-art results in many computer vision tasks~\cite{simonyan2014very, krizhevsky2014imagenet}, we train convolutional neural networks (CNNs) to directly predict the population of a given $0.01\degree \times 0.01\degree$ area using only satellite imagery, then summarize the predictions at different administrative area resolutions. 
These high-level predictions provide greater confidence in the accuracy of our model's predictions at the finer resolution.
We perform two types of model validation. 
Quantitatively, we compare our model's grid cell estimates aggregated at a county level to several US Census county level population projections. 
Qualitatively, we directly interpret the model's predictions in terms of the satellite image inputs.

\section{Related Work}
\label{sec:relatedWork}

Deep learning is being used with increasing frequency to solve problems in the domain of computational sustainability and urban planning.
At a broader level, CNNs have been extensively used in computer vision applications in recent years, and have achieved state of the art results in image classification and object recognition~\cite{krizhevsky2014imagenet, simonyan2014very, he2016deep}. 
New types of network layers, such as batch normalization and dropout, have also been developed to improve the accuracy of CNNs~\cite{ioffe2015batch, srivastava2014dropout}.
Convolutional neural networks have been used to predict the spatial distribution of poverty in developing countries by using nighttime lights as a data rich target for a transfer learning task~\cite{xie2015transfer,jean2016combining}. 
Pre-trained CNNs have recently been shown to be effective at the problem of remote sensing image scenes classification through the tuning a small number of layers~\cite{hu2015transferring, nogueira2017towards}. 
Similarly, deep learning has been shown to be effective in the task of classifying land cover type, with recent work that has achieved high classification accuracy on new large land cover datasets using mixed CNN based approaches~\cite{basu2015deepsat,albert2017using}.

The most similar work to ours also uses CNNs to estimate population from satellite imagery~\cite{doupe2016equitable}. 
The motivation of this paper is similar to ours, as we both attempt to create high-resolution gridded population counts for use in planning applications. 
This paper estimates population in Kenya at a 8km$^2$ resolution with a CNN trained on data from Tanzania at a $250m^2$ satellite pixel resolution. 
The author's propose a way to use their CNN's output as a weighted surface for population disaggregation, and compare this method to other methods for disaggregating population counts in Kenya. 
Our work differs in several important ways. 
First, we focus on validating our model's predictions as raw population projections and do not consider using our model's prediction as a weighted surface for distributing population counts. 
If the population (or projected population) of an area is known a priori, then any population \textit{assignment} method can degrade into a weighting scheme.
Secondly, we focus on interpreting the results of our model as a way of validating its ability to generalize. Thirdly, we apply our method to the entire US using census block derived training and testing data. 

Other related work is divided between the two problems we aim to address jointly with our method: population projection and population disaggregation.
In the following paragraphs we address each of these problems to give context to our methodology.

On average, county population can be reliably extrapolated over short time horizons with simple linear models, however if some counties experience disproportionally higher or lower growth rates, more complicated models are needed~\cite{smith1987tests}. 
The US Census has led research into population and demographic projections, and uses a variety of different population and demographic projection methods to create sub-national projections broken down by age, sex, and race~\cite{long1987survey,long1996postcensal}.
Census postcensal projections, projections done in between census years, are created with a method known as the ratio-correlation method~\cite{schmitt1954accuracy,swanson1994new,long1996postcensal}.
This method uses the current year's estimated population, number of live births, registered vehicles, public school enrollment, registered voters, deaths, and other information to determine the estimated population change at the next census date.
More recently, the American Community Survey has been used as annual supplemental surveys to update the demographics profiles of a variety of sub-national areas in between census years~\cite{mather2005american,censusbureau2009design}.

Population disaggregation methods, and the creation of high resolution population grids have been studied for decades~\cite{deichmann1996review,hay2005accuracy}.
The most basic method in this class is areal interpolation, whereby the known population of an administrative zone is distributed uniformly across its area~\cite{goodchild1993framework}.
This process happens on a discretized grid over an administrative zone, where each cell in the grid is assigned a population value equal to the total population over the total number of cells that cover an administrative zone.
Dasymetric weighting schemes extend this idea of distributing the known population of an area by creating a weighted surface to distribute the known population, instead of doing so uniformly. 
The weighting schemes are determined by combining different spatial layers (e.g., slope, average rainfall, land/water masks) according to some set of rules.
While some weighting schemes are completely ad-hoc, recently, machine learning methods have been used to improve upon this approach~\cite{gaughan2015exploring,sorichetta2015high,stevens2015disaggregating}.
These methodologies are similar to traditional supervised machine learning problems~\cite{linard2011assessing}, but since actual ground truth data does not exist to compare against, validating the results of dasymetric models is challenging. 
Finally, there are many existing gridded population datasets created using a variety of the previously mentioned disaggregation techniques.
Briefly, these include: Gridded Population of the World~\cite{doxsey2015taking}, GRUMP~\cite{schneider2009new}, Landscan~\cite{dobson2000landscan,bhaduri2002landscan}, as well as the AfriPop, AsiaPop, and AmeriPop databases.

\section{Methods}
\label{sec:methods}
The goal of this research is to make high-resolution gridded population estimates from satellite imagery. 
To do this we train CNNs that take satellite imagery of some area as input, and output a population estimate for that area. 
We train our models on the continental United States using US Census population counts and Landsat 7 1-year composite imagery from the year 2000. 
We test our models using the 2010 versions of the same datasets, and evaluate the population estimates in two ways: (1) aggregating our model's estimates at the county geography level, then comparing them to projected county population counts; and (2) showing \textit{why} our model makes predictions in terms of input image features.

As described in Section \ref{sec:data}, we let $\mathbf{P}_t$ be a grid of target population values covering the continental United States, $\mathbf{C}_t$ be a grid of target population class values, and $\boldsymbol{\theta}_t$ be a grid of satellite images, where for every target value $P_t^{i,j}$ and $C_t^{i,j}$ there is an associated satellite image, $\theta_t^{i,j}$. 
Using this notation, we can express our learning task as estimating two functions: one in a regression format, $f(\theta_t^{i,j}) = P_t^{i,j}$, and one in a classification format, $g(\theta_t^{i,j}) = C_t^{i,j}$. 
For the purpose of this study we will focus on the classification version of this problem. 
We use CNNs to approximate this function, as the mapping from image to population counts will be highly non-linear, noisy, and depend strongly on the semantic content of the input image, e.g., on the quantity and type of buildings visible in an input image.
Once we have approximated $g$ on a training year, i.e. for $t=2000$, we can use it to create population projections for a future year, in which a census has not been taken, but satellite imagery exists for. 
We validate this modeling methodology by training CNNs using data from $\mathbf{C}_{2000}$ and $\boldsymbol{\theta}_{2000}$, then running our model with all of $\boldsymbol{\theta}_{2010}$ to create a predicted population surface for 2010. 
To evaluate our predictions, we compare our predicted population values aggregated at the county level to other county level population predictions, we show the errors our models makes, and we use interpretation techniques to uncover why our models are making such predictions.

We describe the data and the preprocessing steps that we use in Section \ref{sec:data}, the CNN model architecture choices in Section \ref{sec:modelArchitecture}, and the experimental methodology that we follow to train, validate, and test our models in Section \ref{sec:experimentalSetup}.
Note that we perform all model training, testing, and experiments using a single desktop workstation containing an NVIDIA Titan GPU.

\subsection{Data} \label{sec:data}

We use three datasets in this work: the Center for International Earth Science Information Networks' (CIESIN) US Census Summary Grids for 2000 and 2010~\cite{census2000,census2010}, Landsat 7 1-year composite images for 2000 and 2010 (courtesy of the U.S. Geological Survey)\footnote{Landsat: \url{https://landsat.usgs.gov/}} downloaded from Google Earth Engine, and county level population data for 2000 and 2010 from the US Census.

The US Census Summary Grids are raster files with a resolution of 30 arc-seconds ($\approx 1km^2$) where the raster cell values are population counts from their respective census.
The per cell counts are created by disaggregating census survey data from census block geographies, while taking into account various geographic features, such as bodies of water, where people won't be living.
In general, a raster cell will contain an area-weighted combination of the populations from the census block shapes that it intersects with.
Since census block geographies are smaller than the 30 arc-second grid in heavily populated areas, these maps represent the closest ``ground truth'' values for population that are available to use as training data for our machine learning models. As a pre-processing step, we re-project these two rasters into a slightly coarser grid with a resolution of $0.01\degree \times 0.01\degree$ ($\approx 1105m^2$ at the equator), where the northwest corner is at $124.849\degree W, 49.3844\degree N$.

We represent each of these grids as a matrix, $\mathbf{P}_t \in \mathbb{Z}_{+}^{2499 \times 5796}$, where an entry $P_t^{i,j}$ represents the population of the cell in the $i^{th}$ row and $j^{th}$ column from year $t$ (in this case $t \in \{2000, 2010\}$).
We further pre-process the data by creating an additional, binned version of each population raster, where a cell takes on a value representing which bin its population count falls in.
Specifically, we create matrices $\mathbf{C}_t$, where an entry $C_t^{i,j} = 0$ if $0 \leq P_t^{i,j} < 1$, $1$ if $2^{1} \leq P_t^{i,j} < 2^{2}$, ..., $k$ if $2^{k} \leq P_t^{i,j} < 2^{k+1}$ where $k\in\mathbb{N}$.
This process discretizes the target population values which simplifies our learning tasks by creating a classification problem. For $\mathbf{C}_{2000}$ the highest class value is $k=17$, representing a cell that has a population in the range $[65,536, 131,072)$.
For the rest of the study, we will use these \textit{population class values} instead of the raw population count values when discussing estimating population.

Landsat 7 1-year composite data is available through Google Earth Engine for the years of 1999 through 2014\footnote{Google Earth Engine: 	\url{https://earthengine.google.com/}}.
The 1-year composites are made by taking the median pixel values from a sample of the least cloudy images from the given year. 
We use data from the 2000 and 2010 sets, with bands 1 through 7, at a $15m^2$ resolution.
This data is downsampled from the native resolution of $30m^2$ recorded by the Landsat 7 satellite using nearest neighbor interpolation.
As a pre-processing step, for every $0.01\degree \times 0.01\degree$ cell in the population matrices, we take the grid of Landsat imagery that it covers.
We resize the grid of Landsat imagery covered by a single population cell into a square volume with a height and width of $74$ pixels, as the number of actual satellite imagery pixels that cover a $0.01\degree \times 0.01\degree$ area will vary with latitude. 
We choose a height and width of 74, because at a latitude of $45\degree N$ (approximately the center of the US), a $0.01\degree \times 0.01\degree$ cell is $\approx1,111m^2$, and with a height and width of $74$ pixels of 15x15 meters, our satellite images will represent a similarly sized $1,110m^2$ area. 
We let the grids of Landsat images be represented as $\boldsymbol{\theta}_{t}$, where by for every $P_t^{i,j}$ cell from the population matrices, we have an associated satellite image volume, $\theta_t^{i,j} \in \mathbb{Z}_{+}^{74 \times 74 \times 7}$.

The county level population data from the US Census includes the ground truth population values for each county in 2000, and 2010, the postcensal population estimates for each county in 2010, and the ACS 5-year 2006-2010 population estimates for each county in 2010. We use this data evaluate our models' aggregate estimates, and refer to the ground truth 2010 county population counts as ``Actual 2010'' in Section \ref{sec:resultsAndDiscussion}.

\subsection{Model Architecture} \label{sec:modelArchitecture}

\begin{figure}
\centering
\includegraphics[width=1.0\linewidth]{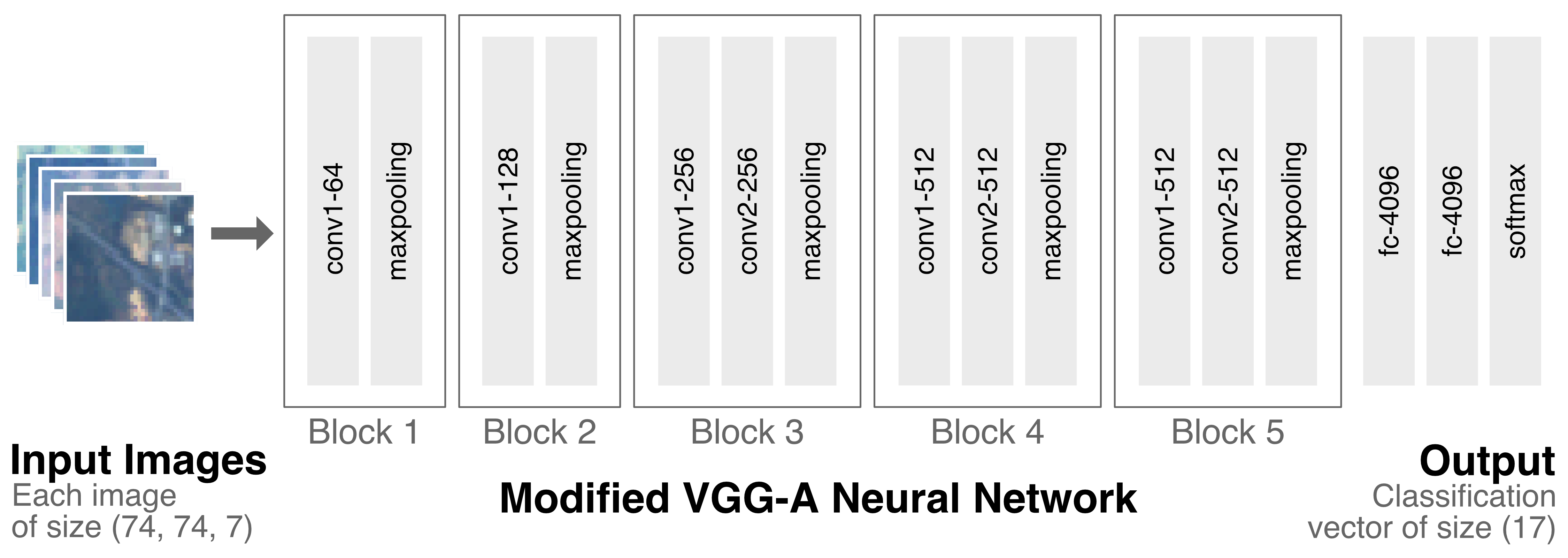}
\caption{Our deep learning model architecture, based off of the VGG-A model. The model inputs satellite images of size (74, 74, 7) in to a linear neural network consisting of 5 convolutional blocks. Each convolutional block contains at least one convolutional layer (conv) and a maxpooling layer. After the 5 convolutional blocks, two fully connected (fc) layers feed into the softmax activated output of length (17) to perform classification.}
\label{fig:arch}
\end{figure}

We experimented with different CNN architectures and hyperparameters using training and validation sets sampled from the 2000 datasets over a $1\degree \times 1\degree$ area in the southeast United States. Our assumption is that a model architecture/hyperparameter set which can perform well on this subset of the entire US will be able to perform equally well throughout the entire study area. The training and validation set sampling was performed through the methodology described in Section \ref{sec:experimentalSetup}.

We considered the 5 well-known `VGG' model architectures, VGG-A through VGG-E from \cite{simonyan2014very}, and variations of each of the 5 VGG architectures that included dropout and batch normalization layers. 
We adapt the VGG architectures to use our input images of size (74,74,7).
Since we have discretized our target values into 17 different classes, we resize the output layer to $17$ and use a softmax activation function.
For all experiments we use a batch size of 512 samples, the Adam optimization method~\cite{kingma2014adam} from the Python Keras library~\cite{chollet2015keras} (with default parameters), the categorical cross entropy loss function, and we train all networks for 30 epochs (with consideration to overfitting through observing the training/validation loss curves). 
We found that a VGG-A architecture results in the best top-1 and top-3 accuracy on both the training and validation sets over 30 training epochs and therefore use this architecture for the remainder of the study. 
See Figure \ref{fig:arch} for a diagram showing the structure of our model.
We chose 30 epochs as a cut off as the best models do not show any improvements in terms of validation loss after this point.

\subsection{Experimental Setup} \label{sec:experimentalSetup}

\begin{figure}
\centering
\raisebox{-.5\height}{\includegraphics[width=0.49\textwidth]{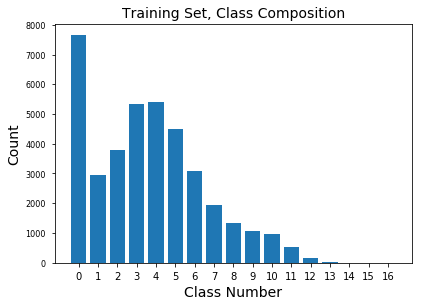}}
\hfill
\raisebox{-.5\height}{\includegraphics[width=0.49\textwidth]{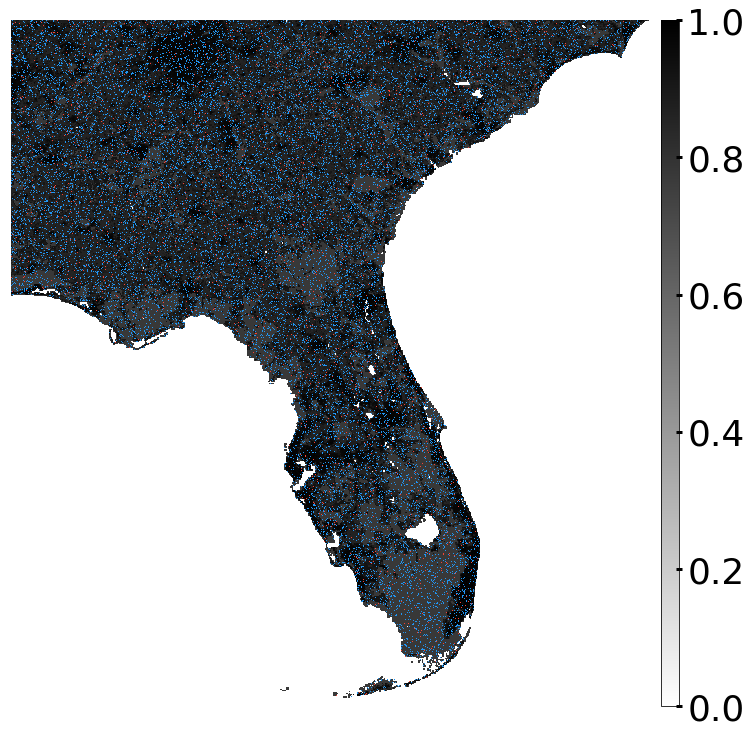}}
\caption{Training/validation set sampling technique. (Left figure) shows the counts of samples in the training set belonging to each of the target classes (i.e. the $C_t^{i,j}$ values). The target class values in the validation set follows the same distribution. (Right figure) shows the probability surface from which the training and validation points are sampled from; samples from the training set (38738 points) are shown in blue, and samples from the testing set (3874 points) are shown in red.}
\label{fig:sampling}
\end{figure}

Our study area consists of a $2,499$ by $5,796$ grid covering the continental United States that contains $\approx 8$ million target values.
As using all of these samples to train with presents a significant computational challenge, we divide up the study area into 15, $1,000$ by $1,000$ ($1\degree \times 1\degree$) \textit{chunks}, and train an independent model for each chunk according to the methods described in Section \ref{sec:modelArchitecture}.
Recent work using random forest models for population mapping suggests that, ``more accurate population maps can be produced by using regionally-parameterized models where more spatially refined data exists''~\cite{gaughan2015exploring}, which we follow with this methodology.
Within each chunk we sample $1/10$th of the available data to use as training samples, and $1/100th$ of the data to use as validation samples. 
As there is a class imbalance problem in the population data, with many more samples in the lower population classes than in the higher population classes, we perform a weighted sampling to select training and validation points.
We let $c_i$ represent the number of points in class $i$ over the entire training set, then the probability of selecting a point $C_t^{i,j} = x$ is given as $1 - c_x / \sum_{i=1}^{17} c_i$. This sampling methodology serves to undersample the higher frequency classes more often than the lower frequency ones, while still resulting in a representative sample of all classes from the study area. 
Figure \ref{fig:sampling} shows the results of this sampling methodology.

An important component of any machine learning or modeling application is validating that the models are able to generalize well to unseen data, and that the models are able to make reasonable predictions. 
It is important to note that because there does not exist any true ``ground truth'' gridded population data, it is not possible to truly evaluate population disaggregation techniques. 
As the purpose of our models is to predict population values from only satellite imagery, they should (a) be able to make reasonable population predictions when compared to other population prediction techniques, (b) be interpretable, where population predictions are able to be explained in terms of semantic features of the input images, and (c) should have explainable errors. 
We address each of these three points in the following three paragraphs.

We first evaluate our results by comparing our model's aggregate population estimates at the county level with US Census Postcensal county level estimates for 2010 (\textbf{POSTCENSAL})~\cite{long1996postcensal}, and American Community Survey 5-year estimates for 2006-2010 (\textbf{ACS5YR})~\cite{censusbureau2009design} in terms of accuracy when evaluated against the actual 2010 Census \cite{census2010}.
We convert our per grid cell population class predictions, $\hat{C}^{i,j}$, into county level population estimates, $\hat{P}^{i,j}$, in two ways. 
The first method (\textbf{CONVRAW}), involves converting the class values directly into population values as described in Equation \ref{equ:predictions}.
\begin{equation} \label{equ:predictions}
\hat{P}^{i,j} = \begin{cases} 
  0 & \hat{C}^{i,j} = 0 \\
   \frac{1}{2} (2^{\hat{C}^{i,j} - 1} + 2^{\hat{C}^{i,j}}) & \text{otherwise}
\end{cases}
\end{equation}
\noindent This formula is equivalent to predicting the middle point of each class bin as the population estimate. 
We sum the predicted population values for each cell whose centroid falls within a particular county to get the aggregate county predictions. 
The second method, (\textbf{CONVAUG}), uses the values from the softmax activations in the last layer of each CNN as ``features'' into a secondary machine learning model. 
Specifically, the last layer of our CNN models has a width of 17, where the output values represent the probability that the input image belongs to each of the 17 population classes. 
We run our CNN models for each cell in the training dataset (covering the entire US), and record the output vector at each location. 
We aggregate the output vectors by county by summing the vectors of all pixels that are covered by each county. 
This process gives us a \textit{feature vector} for each county which contains information about the composition of the population classes of the cells that make up that county. 
We then use these feature vectors to train a gradient boosting model to predict the ground truth county population values from the training set year. 
We perform the same process on the test set to create feature vectors with our trained CNN models and use the trained gradient boosting model to make county level population estimates. 
While this methodology is somewhat orthogonal to the main points of this paper, it shows how our trained CNN models can be used as a mechanism for feature extraction, and that the features the model learns are indeed valid signals of population numbers. 
We show the results from this county level evaluation in Section \ref{sec:expCountyEval}.

As described in the previous paragraph, for each input cell our model outputs a probability distribution over the possible population class values. 
Using this, we create maps that show the probability that each cell belongs to a given class. 
Similarly, we show which input images maximally activate every given output class. 
We show these interpretability results in Section \ref{sec:expInterpret}

Finally, we interpret the largest errors that our model makes. 
Because our model is limited to using satellite imagery data, it will become ``confused'' in cases where there are signs of human settlements that do not manifest as populated in the census datasets. 
This confusion is evidence that our models are able to learn the higher-order features as to what constitutes ``populated areas'', however do not have enough data to discriminate between different types of human activities. 
The results and discussion of this are shown in Section \ref{sec:expErrors}.

\section{Results and Discussion}
\label{sec:resultsAndDiscussion}

Our results focus on validating the modeling methodology, and are broken down into three sections: evaluating how good our model's population estimates are when aggregated at the county level in Section \ref{sec:expCountyEval}, interpreting why our models make the predictions that they do in \ref{sec:expInterpret}, and evaluating and explaining our model's per pixel errors when compared with ground truth in Section \ref{sec:expErrors}. 

\subsection{County level Estimates} \label{sec:expCountyEval}

\begin{figure}[!b]
\centering
\includegraphics[width=0.78\textwidth]{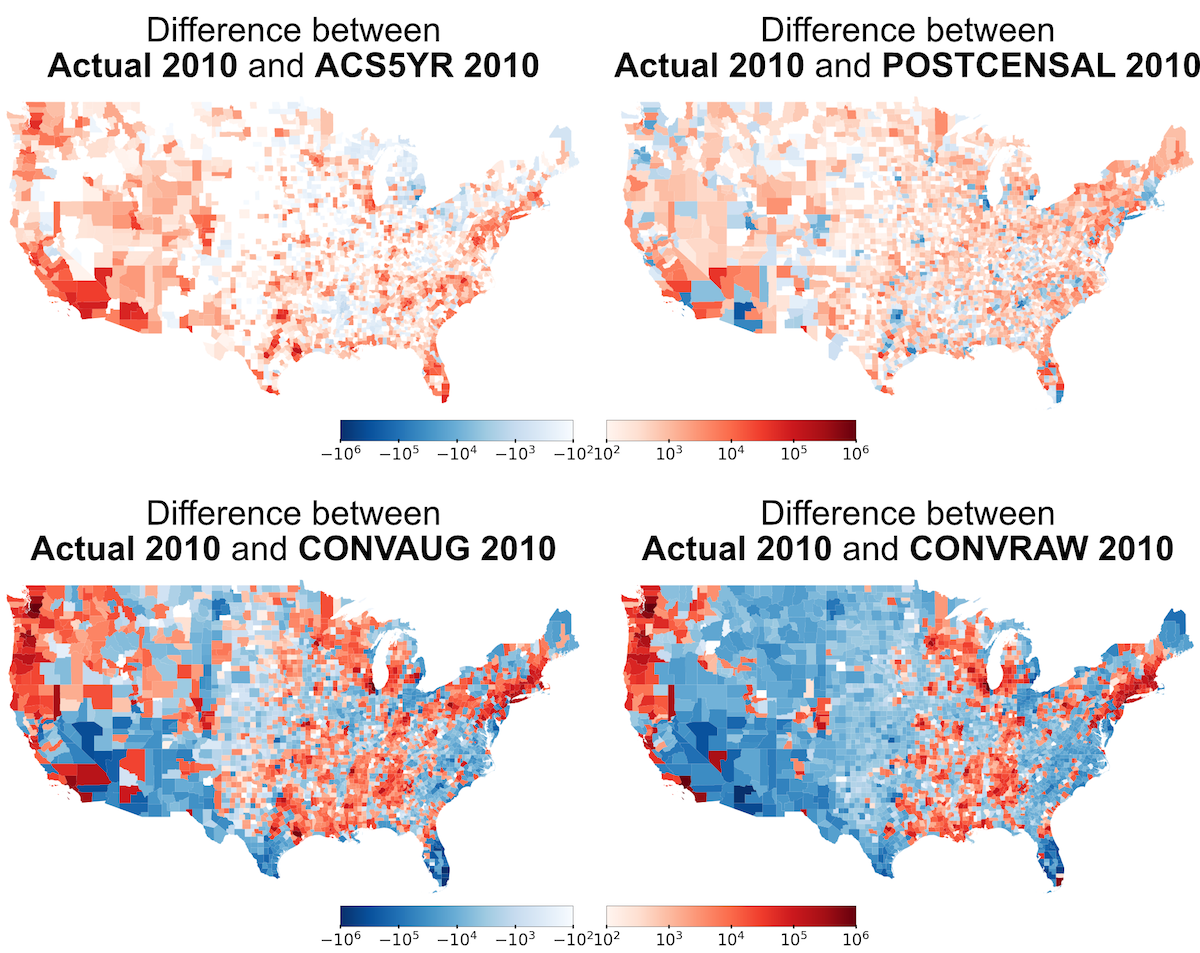}
\caption{County level population projection results. Difference between the ground truth 2010 county population values and the tested methods for estimating county populations.}
\label{fig:countyResults}
\end{figure}

Here we compare 4 different methods for predicting county level population counts for the continental US in 2010. 
The four methods are as described in Section \ref{sec:experimentalSetup}: \textbf{POSTCENSAL}, \textbf{ACS5YR}, \textbf{CONVRAW}, and \textbf{CONVAUG}.
None of these methods contain information about the true population counts for the target year, 2010, therefore must infer the population either from detailed historical population and demographic data in the case of \textbf{POSTCENSAL}, supplemental survey information in the case of \textbf{ACS5YEAR}, or a combination of satellite and historical population  data in the case of our methods \textbf{CONVRAW} and \textbf{CONVAUG}.
We compare the predicted populations for all counties with each method to the ground truth population taken from the US 2010 Census and record the mean absolute error (Mean AE), median absolute error (Median AE), $r^2$ score, and mean absolute percentage error (MAPE).
The results for this comparison can be found in Table \ref{tbl:countyResults}, and the per county errors for each method are visualized in Figure \ref{fig:countyResults}.

\begin{table}[t]
\centering
\resizebox{0.6\textwidth}{!}{%
\begin{tabular}{@{}lrrrr@{}}
\toprule
 & \multicolumn{1}{l}{Mean AE} & \multicolumn{1}{l}{Median AE} & \multicolumn{1}{l}{$r^2$} & \multicolumn{1}{l}{MAPE} \\ \midrule
CONVRAW & 23,005 & 6,357 & 0.9103 & 73.78 \\
CONVAUG & 19,484 & 4,642 & 0.9365 & 49.82 \\
POSTCENSAL & 2,020 & 559 & 0.9993 & 3.09 \\
ACS5YR & 1,704 & 214 & 0.9996 & 34.44 \\ \bottomrule
\end{tabular}%
}
\caption{County level population projection results. Comparison of 4 techniques for estimating 2010 county population for all counties in the continental United States.}
\label{tbl:countyResults}
\end{table}

The two statistical methods used by the US Census provide more accurate predictions of county level population for 2010, and have lower median and mean absolute errors than our two methods. 
This result is expected, as the predictions made by these methods take many more historical features into account, while our methods only use the previous census' population counts and satellite imagery to make predictions. 
Our model's mean and median errors fall within an order of magnitude of the census model's errors, and our model's MAPE is similar to the ACS5YR results. 
We perform this comparison to validate that our model's unaided population estimates are not wildly off, which suggests that our model is able to capture the true signal in determining population values from satellite imagery. 
Considering the evaluation of how well our model captures the \textit{locations} of populations, we argue that because our aggregate estimates at the county level are not wildly off, our model's individual cell predictions must be approximately valid.
Similar to population disaggregation methodology, our model's individual cell predictions will be the most accurate when they are scaled to match the true population value, or a trusted population estimate. 
While these county level estimates should not be used in place of the more accurate census estimation methods in the US, they could be used to create continuously updated population maps for developing countries that do not have the detailed data required to run population projection models.

\subsection{Prediction Interpretability} \label{sec:expInterpret}

\begin{figure}[!b]
\centering
\includegraphics[width=1.0\linewidth]{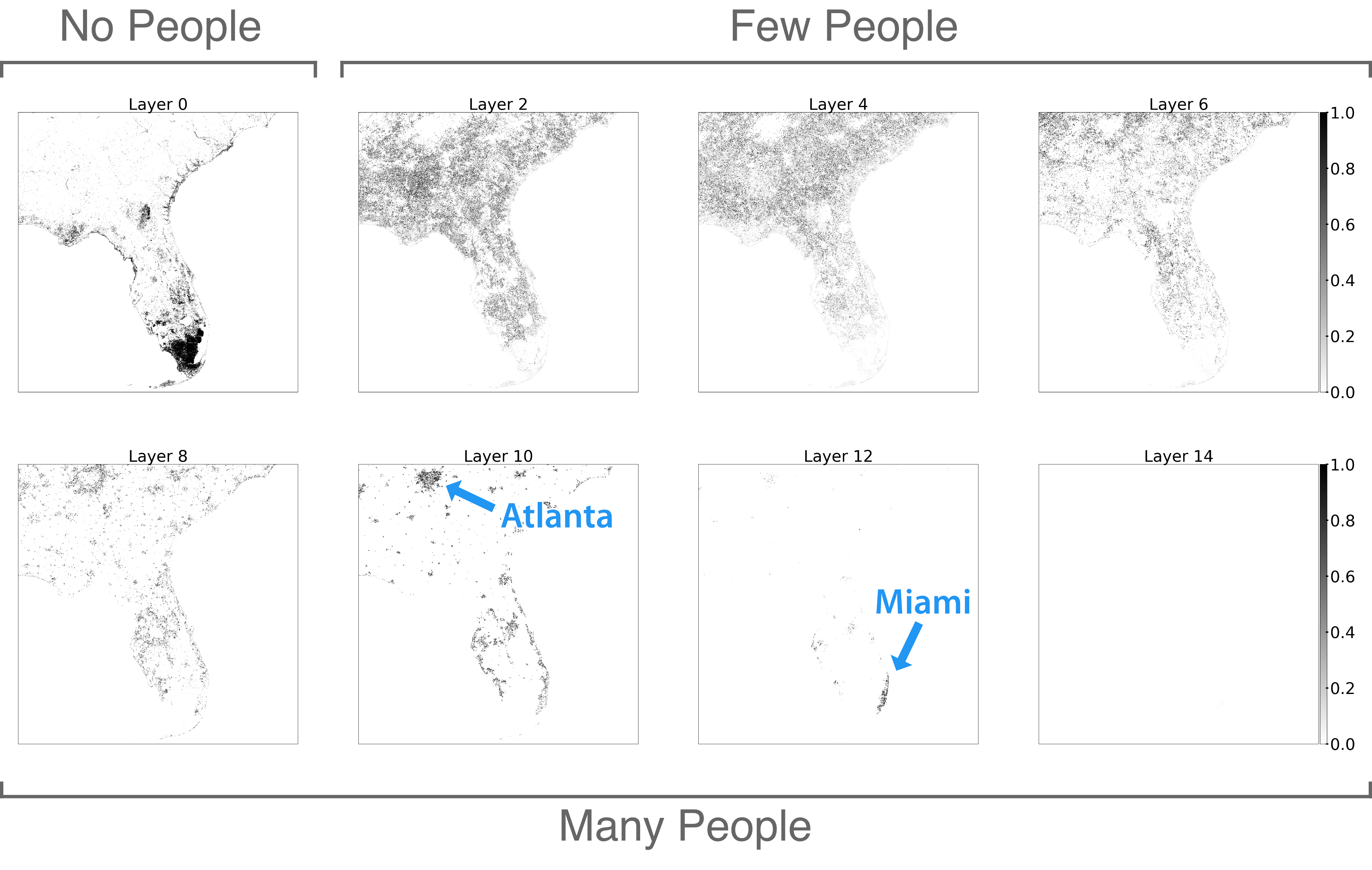}
\caption{Activation maps for eight different population classes on the southeastern United States. Each map shows the estimated probability that a cell belongs in the map's population class. Layer 0 corresponds to zero people, layers 2, 4, and 6 correspond to few people, and layers 8, 10, 12, and 14 correspond to many people living in the activated areas. Notice the higher the layer number the more dense the population becomes, which naturally highlights urban cities such as Atlanta and Miami, annotated above.}
\label{fig:activationMaps}
\end{figure}

\begin{figure}
\centering
\includegraphics[width=0.8\textwidth]{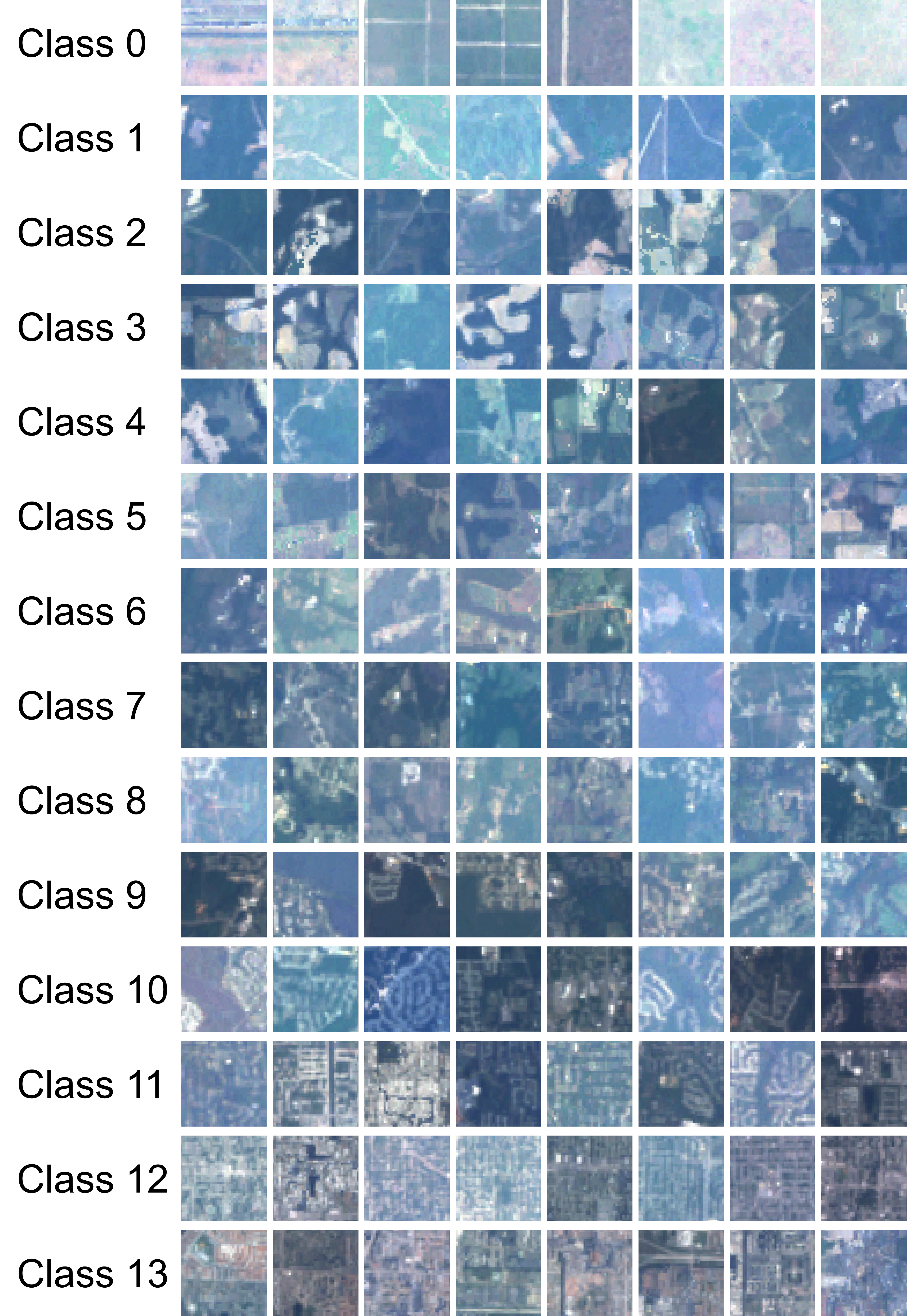}
\caption{The top 8 most confident prediction images from the test set for each class (e.g. 99\% prediction for a given class), all of which are correctly classified. Notice the types of images that appear from top (highways, few people) to bottom (buildings, many people) further indicated that our deep learning model is learning semantically-relevant features from satellite imagery.}
\label{fig:classRepresentatives}
\end{figure}

Interpretability is an important aspect of any modeling process. 
As we cover in Section \ref{sec:relatedWork}, some population disaggregation methods rely on ad-hoc rules to assign the population of an administrative area to the grid cells that cover the same area. 
In some applications, the methods for determining these rules, or the rules themselves, are available, while in other products, such as Landsat~\cite{dobson2000landscan,bhaduri2002landscan}, the methodology is not public, and therefore, subsequent years of predictions are not comparable.
Additionally, while some basic dasymetric heuristics, such as ``humans do not live on land where the slope is over $45\degree$'', can be globally applied, more detailed heuristics might be region specific. 
Our methodology bypasses these potential problems by generating the probability that a section of satellite imagery belongs to each population class, which allows us to show how confident our models are about a certain classification decision.
Similarly, because our model only considers satellite imagery as input, all of the predictions made by our model will be able to be explained in terms of the features of the input image.
We show these two components of our methodology in Figures \ref{fig:activationMaps} and \ref{fig:classRepresentatives} respectively.
\\
\\
In Figure \ref{fig:activationMaps} we show maps for several of the output population classes that show the estimated probability of each pixel belonging to the respective class. 
From these we observe that our model makes confident predictions about the 0 population class (Layer 0), and the higher population classes. 
The lack of confidence in the lower population classes (Layers 2 and 4) makes sense as we do not expect the visual difference between $1$km$^2$ areas in which 4 and 16 people live to be large. 
To compound this, census block geographies are larger in low population rural areas, meaning that our disaggregated ``ground truth'' training data will be noisier in lower population areas.
In Figure \ref{fig:classRepresentatives} we show, for each class, the top 8 satellite image inputs from the testing set, that maximize the softmax output for that class. 
These images give us an insight into what types of features our model is learning. 
There are clear patterns moving from the lower classes, which represent sparsely populated areas, to very the upper classes which represent more urbanized areas. 
In the lower classes, most of the images contain some sort of roadway or distinctively marked fields. 
In classes 6 through 9 there are several buildings and developments visible, while finally in classes 10 through 14 there are dense suburban and urban developments with gridded patterns visible. 

\subsection{Prediction Errors} \label{sec:expErrors}

\begin{figure}[!b]
\centering
\includegraphics[width=1.0\linewidth]{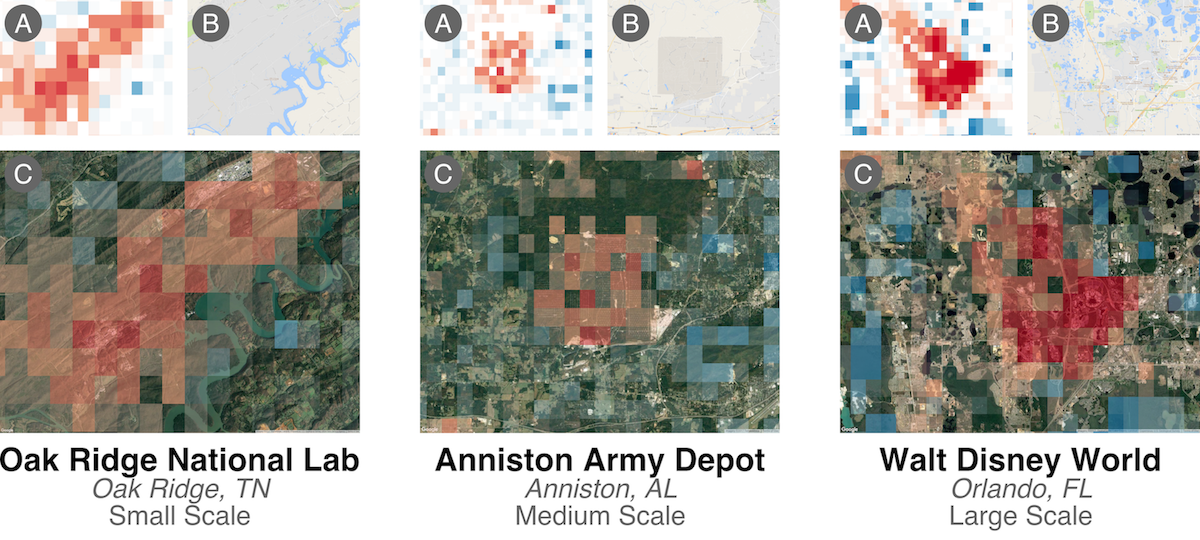}
\caption{Three regions that have particularly high class prediction errors. Red pixels are over-predictions; blue pixels are under predictions. Upon inspection, these three regions are large-scale human-made areas that contain features typically associated with high-population areas, but in reality have very few people living in them. These include Oak Ridge National Lab (left, smaller scale), Anniston Army Depot (middle, medium scale), and Walt Disney World (right, large scale). (A) shows the class prediction errors, (B) shows the same region from Google Maps, and (C) shows (A) overlaid on the satellite imagery.}
\label{fig:modelErrors}
\end{figure}

Here we show some of the errors of our model. 
Through inspecting the pixel class errors, i.e., the true population class value in 2010 (disaggregated from the Census population counts) minus the predicted population class values, we noticed that our model is systematically over-predicting some large areas. 
In Figure \ref{fig:modelErrors} we show three of these cases: Oak Ridge National Laboratory in Oak Ridge, TN, Anniston Army Depot in Anniston, AL, and Walt Disney World in Orlando, FL. 
These locations all share the property of having many man-made structures and signals of human activity, without the ``ground truth'' labeling of a population count from the Census data. 
Walt Disney World has many structures that look similar to those in high population residential areas, and therefore will always be mis-classified by a model that only relies on satellite imagery as input. 
In these cases, a traditional dasymetric modeling approach to disaggregating population will have an advantage over our model, as such an augmented approach could easily incorporate layers describing army bases, amusement parks, and other large spatial structures that will \textit{not} have populations living within their borders. 
Finally, these observations are further evidence that our model is generalizing and learning useful semantic content about the input images with which to make its prediction.

\section{Future Work and Conclusion}
\label{sec:conclusion}

Our goal in this work is to train convolutional neural networks to create high-resolution gridded population maps using only satellite imagery, then validate our model's predictions both quantitatively and qualitatively. 
We predict population counts in the continental US at a $0.01\degree \times 0.01\degree$ ($\approx 1$km$^2$) resolution for 2010, after training on data from 2000. 
To evaluate and validate our models, we first aggregate the population predictions at the county level, and compare them to ground truth county population counts from the 2010 census. 
Our models perform well on the task of projecting county population, with the best model having a median absolute error of 4,642, and although they are not better than traditional county population projection methods used by the US Census, they are able to make reasonable predictions. 
Secondly, we show what the models have learned by creating maps that show the estimated probability of each cell belonging to a given class, and by visualizing the satellite image inputs for each class that our model is most confidently classifying. 
We observe that the most confident images for each class follow an expected pattern, whereby images of rural areas with small roads and fields are classified as low population cells, and gridded urban areas with dense housing are classified as high population cells. 
Finally we qualitatively explain some of the errors that our model is making in terms of noisy input data; for example, our model predicts that an army base in Anniston, Alabama is a high population area, even though the ``ground truth'' census data says that the area is unpopulated.

For future work we plan on extending our current methodology in several different ways. 
In terms of the CNN training process, there are several changes and experiments that we would like to try: experimenting with different loss functions and loss function weighting schemes that could take the ordinal nature of our classification problem into account.
Currently we optimize the categorical cross entropy, which will not discriminate between ``small'' and ``large'' errors, i.e., the loss will not penalize misclassifying a label with true class 11, as a 10, more than it would penalize misclassifying the 11 as a 1. 
We also would like to try training a model on the entire US; as this task has the potential to use over 8 million samples, this will bring entirely different challenges to the deep learning process. 
In terms of applying and evaluating the models, we would like to use these models to predict population counts in countries where censuses are not taken as often, and are not taken at as fine of a resolution as in the US. 
Similarly, we want to experiment with the trade-offs between ground truth data resolution and model accuracy to determine the limits of the applicability of these models. 
Finally, we would like to apply transfer learning methods to this problem such as investigating whether pre-training models on land-use classification tasks result in better predictions or whether directly predicting nighttime light intensities helps.

\bibliographystyle{unsrt}
\bibliography{citations} 

\end{document}